\documentclass{article}
\usepackage{ijcai21}

\usepackage{times}
\usepackage{soul}

\usepackage[utf8]{inputenc}
\usepackage[small]{caption}
\usepackage{graphicx}

\usepackage{booktabs}
\usepackage{url}
\usepackage{amsmath,amsfonts,bm}

\def\eqref#1{equation~\ref{#1}}

\def\1{\bm{1}}

\def\eps{{\epsilon}}

\DeclareMathAlphabet{\mathsfit}{\encodingdefault}{\sfdefault}{m}{sl}
\SetMathAlphabet{\mathsfit}{bold}{\encodingdefault}{\sfdefault}{bx}{n}

\DeclareMathOperator*{\argmax}{arg\,max}

\usepackage[hidelinks]{hyperref}

\usepackage[ruled, vlined]{algorithm2e}
\usepackage{diagbox}
\usepackage{slashbox}
\usepackage{multirow}
\usepackage[caption=false]{subfig}
\usepackage{xcolor}
\usepackage{subfiles}
\usepackage{amsmath}
\usepackage{amsthm}
\usepackage{graphicx}

\usepackage{amsfonts,amssymb}
\theoremstyle{definition}
\newtheorem{definition}{Definition}
\newtheorem{theorem}{Theorem}
\newtheorem{lemma}[theorem]{Lemma}

\newcommand{\mb}[1]{\mathbf{#1}}
\newcommand{\tb}[1]{\textbf{#1}}
\newcommand{\mc}[1]{\mathcal{#1}}

\newcommand{\M}{\mathcal{M}}
\newcommand{\D}{\mathcal{D}}
\newcommand{\eqdef}{\stackrel{\Delta}=}
\newcommand{\x}{\mathbf{x}}
\newcommand{\aux}{\textsf{aux}}
\newcommand{\del}[1]{}
\makeatletter
\newcommand*{\rom}[1]{\expandafter\@slowromancap\romannumeral #1@}
\makeatother

\newcommand{\rev}[1]{\textcolor{black}{#1}}
\newcommand{\HBS}[1]{\textcolor{black}{#1}}

\title{Practical One-Shot Federated Learning for Cross-Silo Setting}

\author{
Qinbin Li$^1$
\and
Bingsheng He$^1$\and
Dawn Song$^2$
\affiliations
$^1$National University of Singapore\\
$^2$University of California, Berkeley
\emails
\{qinbin, hebs\}@comp.nus.edu.sg,
dawnsong@cs.berkeley.edu
}

\begin{document}

\maketitle

\begin{abstract}
Federated learning enables multiple parties to collaboratively learn a model without exchanging their data. While most existing federated learning algorithms need many rounds to converge, one-shot federated learning (i.e., federated learning with a single communication round) is a promising approach to make federated learning applicable in cross-silo setting in practice. However, existing one-shot algorithms only support specific models and do not provide any privacy guarantees, which significantly limit the applications in practice. In this paper, we propose a practical one-shot federated learning algorithm named FedKT. By utilizing the knowledge transfer technique, FedKT can be applied to any classification models and can flexibly achieve differential privacy guarantees. Our experiments on various tasks show that FedKT can significantly outperform the other state-of-the-art federated learning algorithms with a single communication round. 

\end{abstract}

\section{Introduction}

While the size of training data can influence the machine learning model quality a lot, the data are often dispersed over different parties in reality. Due to regulations on data privacy, the data cannot be centralized to a single party for training. A popular solution is federated learning \cite{kairouz2019advances,li2019survey,yang2019federated}, which enables multiple parties to collaboratively learn a model without exchanging their local data.

A typical and widely used federated learning algorithm is FedAvg \cite{mcmahan2016communication}. Its training is an iterative process with four steps in each iteration. First, the server sends the global model to the selected parties. Second, each of the selected parties updates its model with their local data. Third, the updated models are sent to the server. Last, the server averages all the received models to update the global model. There are many variants of FedAvg \cite{li2018federated,karimireddy2019scaffold,lin2020ensemble}, which have similar frameworks to FedAvg.

The above iterative algorithms are mainly designed for the cross-device setting, where the parties are mobile devices. In such a setting, the server is usually managed by the federated learning service provider (e.g., Google) and the parties are the users that are willing to improve their service quality (e.g., Google Keyboard users). The server can sample part of devices to conduct federated learning in each round and there are always users available. However, in cross-silo settings, where the parties are usually organizations, approaches like FedAvg may not work in practice due to the following reasons. First, the algorithm requires parties to participate multi-round training, which is not practical in some scenarios such as model markets \cite{vartak2016modeldb,baylor2017tfx}. Second, federated learning across rounds may suffer from attacks (e.g., inference attacks \cite{shokri2017membership}) from curious parties. Last, it is hard to find a fair and trusted server to lead the training process.   

A promising solution is one-shot federated learning (i.e., federated learning with only a single communication round). With one-shot federated learning, the parties can simply sell or upload their local models to a model market (or a model management platform) \cite{vartak2016modeldb,baylor2017tfx}. Then, a buyer or the market can use these models \HBS{collectively} to learn a final model\HBS{, which is very suitable for cross-silo settings}. \HBS{Such a process largely reduces the multi-round requirements on the stability of the parties. It is natural that sellers put their models into the model market (in return for incentives and benefits, whose design is interesting but out of scope of this paper). Usually, sellers are not engaged in purchasing and consuming the models, and one-shot federated learning algorithms are a must here.}

There have been several studies on one-shot federated learning \cite{guha2019one,zhou2020distilled,pnfm}. However, existing one-shot federated learning studies have the following obvious shortcomings. First, they usually are specially designed for a special model architecture (i.e., support vector machines or multi-layer perceptrons), which significantly limit the applications in the real-world. Second, they do not provide any privacy guarantees. This is important in the model market scenario, since the models may be sold to anyone including attackers. 

In this paper, we propose a new one-shot federated learning algorithm named FedKT (Federated learning via Knowledge Transfer). Inspired by the success of the usage of unlabelled public data in many studies \cite{papernot2016semi,papernot2018scalable,chang2019cronus,lin2020ensemble}, which often exists such as text and images and can be obtained by public repositories or synthetic data generator or various data markets , we design a two-tier knowledge transfer framework to achieve effective and private one-shot federated learning.  As such, unlike most existing studies that only work on either differentiable models (e.g., neural networks \cite{mcmahan2016communication}) or non-differentiable models (e.g., decision trees \cite{li2020practical}), FedKT is able to learn any classification model. Moreover, we develop differentially private versions and theoretically analyze the privacy loss of FedKT in order to provide different differential privacy guarantees. Our experiments on various tasks and models show that FedKT significantly outperforms the other state-of-the-art federated learning algorithms with a single communication round.

\rev{Our main contributions are as follows.
\begin{itemize}
    \item Based on the knowledge transfer approach, we propose a new federated learning algorithm named FedKT. To the best of our knowledge, FedKT is the first one-shot federated learning algorithm which can be applied to any classification models.
    \item We consider comprehensive privacy requirements and show that FedKT is easy to achieve both example-level and party-level differential privacy and theoretically analyze the bound of its privacy cost.
    \item We conduct experiments on various models and tasks and show that FedKT can achieve much better accuracy compared with the other federated learning algorithms with a single communication round.
\end{itemize}
}

\section{Background and Related Work}


\subsection{Knowledge Transfer}

Knowledge transfer has been successfully used in previous studies \cite{hinton2015distilling,papernot2016semi,papernot2018scalable}. Through knowledge transfer, an ensemble of models can be compressed into a single model. A typical example is the PATE (Private Aggregation of Teacher Ensembles) \cite{papernot2016semi} framework. In this framework, PATE first divides the original dataset into multiple disjoint subsets. A teacher model is trained separately on each subset. Then, the max voting method is used to make predictions on the public unlabelled datasets with the teacher ensemble, i.e., choosing the majority class among the teachers as the label. Last, a student model is trained on the public dataset. A good feature of PATE is that it can easily satisfy differential privacy guarantees by adding noises to the vote counts. Moreover, PATE can be applied to any classification model regardless of the training algorithm. However, PATE is not designed for federated learning. 


\subsection{\rev{Federated Learning with Knowledge Transfer}}

There have been several studies \cite{li2019fedmd,chang2019cronus,he2020group,lin2020ensemble,zhu2021votingbased} using knowledge transfer in federated learning. \cite{li2019fedmd} needs a public labeled dataset to conduct initial transfer learning, while FedKT only needs a public unlabeled dataset. \cite{chang2019cronus} and \cite{he2020group} have different objectives from FedKT. Specifically, \cite{chang2019cronus} designs a robust federated learning algorithm to protect against poisoning attacks. \cite{he2020group} considers cross-device setting with limited computation resources and uses group knowledge transfer to reduce the overload of each edge device. \cite{lin2020ensemble} has a similar setting with FedKT. They use a public dataset to improve the global model in the server side. As we will show in the experiment, FedKT has a much better accuracy than \cite{lin2020ensemble} with a single round.

All the above studies conduct in an iterative way, which require many communication rounds to converge and cannot be applied in the model market scenario. Moreover, all existing studies transfer the prediction vectors (i.g., logits) on the public dataset between clients and the server. As we will show in Section \ref{sec:fedkt}, FedKT transfers the voting counts and can easily satisfy differential privacy guarantees with a tight theoretical bound on the privacy loss.

We notice that there is a contemporary work \cite{zhu2021votingbased} which also utilizes PATE in federated learning. While they simply extend PATE to a federated setting, we design a two-tier PATE structure and provide more flexible differential privacy guarantees.

\subsection{One-Shot Federated Learning}
There have been several studies \cite{pnfm,guha2019one,zhou2020distilled,kasturi2020fusion} on one-shot federated learning. Instead of simply averaging all the model weights in FedAvg, \cite{pnfm} propose PFNM by adopting a Bayesian nonparametric model to aggregate the local models when they are multilayer perceptrons (MLPs). Their method shows a good performance in a single communication round and can also be applied in multiple communication rounds. \cite{guha2019one} propose an one-shot federated learning algorithm to train support vector machines (SVMs) in both supervised and semi-supervised settings. \cite{zhou2020distilled} and \cite{kasturi2020fusion} transfer the synthetic data or data distribution to the server, which trains the final model using generated dataset. Such data sharing approaches do not fit the mainstream model sharing schemes. Moreover, all existing one-shot federated learning studies do not provide privacy guarantees, which is important especially in the model market scenario where the models are sold and may be bought by anyone including attackers.


\subsection{Differential Privacy}
Differential privacy~\cite{dwork2014algorithmic} is a popular standard of privacy protection. It guarantees that the probability of producing a given output does not depend much on whether a particular data record is included in the input dataset or not. It has been widely used to protect the machine learning models \cite{abadi2016deep,choquette-choo2021capc}.

\begin{definition} ($(\varepsilon, \delta)$-Differential Privacy)
Let $\M \colon \D \rightarrow \mc{R}$ be a randomized mechanism with domain $\D$ and range $R$. $M$ satisifes  $(\eps,\delta)$-differential privacy if for any two adjacent inputs $d, d'\in \D$ and any subset of outputs $S\subseteq \mc{R}$ it holds that: 
	\begin{equation}
	\label{eq:dp}
	\Pr[\M(d)\in S]\leq e^{\eps}\Pr[\M(d')\in S]+\delta.
	\end{equation}

\end{definition}

The moments accountant method~\cite{abadi2016deep} is a state-of-the-art approach to track the privacy loss. We briefly introduce the key concept, and refer readers to the previous paper~\cite{abadi2016deep} for more details.

\begin{definition} (Privacy Loss)
Let $\M \colon \D \rightarrow \mc{R}$ be a randomized mechanism. Let \textsf{aux} denote an auxiliary input. For two adjacent inputs $d, d' \in \D$, an outcome $o \in \mc{R}$, the privacy loss at $o$ is defined as:
\begin{equation}
  c(o; \M,  \textsf{aux}, d, d') \eqdef \log \frac{\Pr[\M( \textsf{aux}, d) = o]}{\Pr[\M( \textsf{aux}, d') = o]}.
\end{equation}
\end{definition}

\begin{definition} (Moments Accountant)
	Let $\M \colon \D \rightarrow \mc{R}$ be a randomized mechanism. Let \textsf{aux} denote an auxiliary input. For two adjacent inputs $d, d'$, the moments accountant is defined as:
	\begin{equation}\label{eq:moments-accountant}
	\alpha_\M(\lambda) \eqdef \max_{\textsf{aux}, d, d'} \alpha_\M(\lambda;\textsf{aux},d,d')
	\end{equation}
	where $\alpha_\M(\lambda;\textsf{aux},d,d') \eqdef 
	\log \mathbb{E}_o[\exp(\lambda c(o; \M, \textsf{aux}, d, d'))]$ is the log of moment generating function.
\end{definition}

The moments have good composability and can be easily converted to $(\varepsilon, \delta)$-differential privacy \cite{abadi2016deep}. 

\paragraph{Party-level Differential Privacy}
In addition to the standard example-level differential privacy, party-level differential privacy \cite{geyer2017differentially,brendan2018learning} is more strict and attractive in the federated setting. Instead of aiming to protect a single record, party-level differential privacy ensures that the model does not reveal whether a party participated in federated learning or not. 

\begin{definition} (Party-adjacent Datasets)
Let $d$ and $d'$ be two datasets of training examples, where each example is associated with a party. Then, $d$ and $d'$ are party-adjacent if $d'$ can be formed by changing the examples associated with a single party from $d$.

\end{definition}
\section{Our Approach}
\label{sec:fedkt}

\begin{figure}[!]
\centering
\includegraphics[width=\columnwidth]{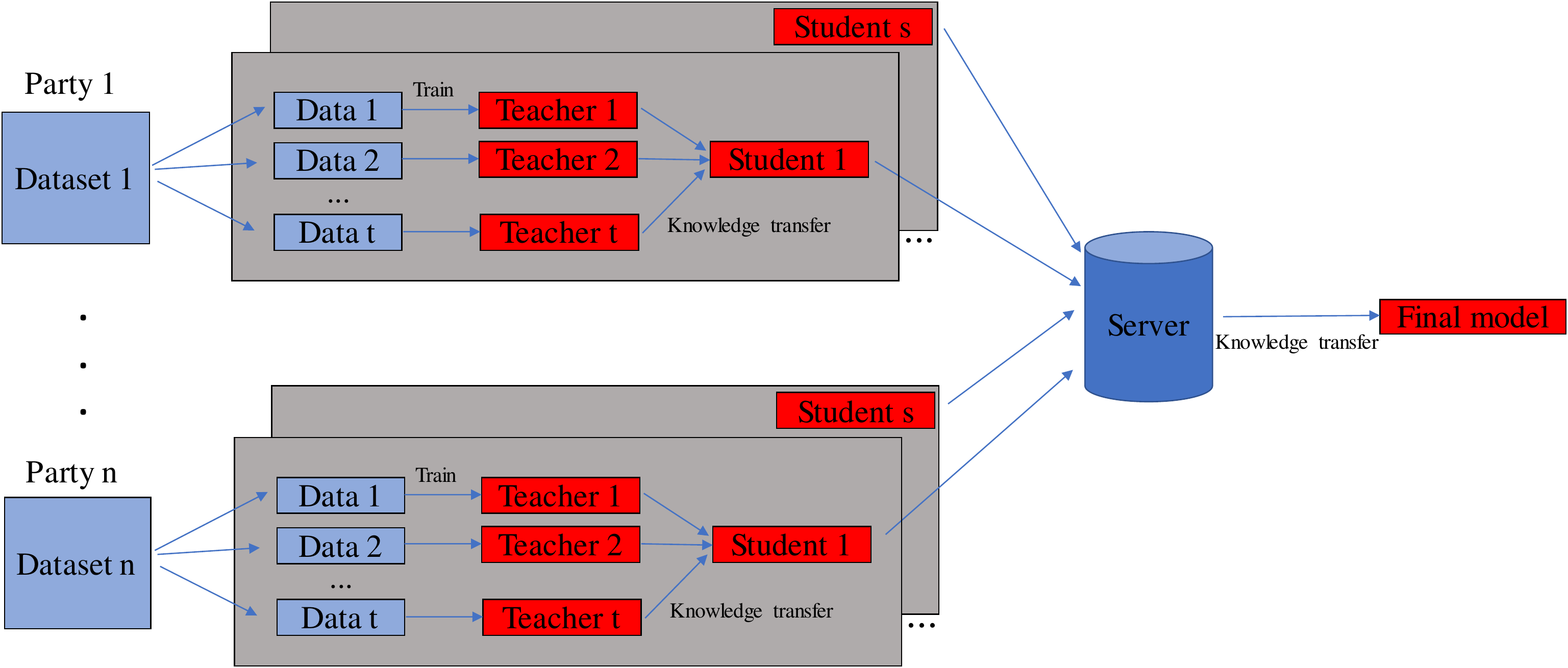}%
\caption{The framework of FedKT}
\label{fig:framwork}
\end{figure}

\paragraph{Problem Statement} Suppose there are $n$ parties $P_1, .., P_n$. We use $\D^i$ to denote the dataset of $P_i$. With the help of a central server and a public unlabelled dataset $\D_{aux}$, our objective is to build a machine learning model over the datasets $\bigcup_{i\in[n]} \D^i$ without exchanging the raw data. Moreover, the learning process should be able to support three different privacy level settings: (1) $L0$: like most federated learning studies \cite{mcmahan2016communication,pnfm,lin2020ensemble}, the possible inference attacks \cite{shokri2017membership,fredrikson2015model} on the models are not considered. \HBS{At $L0$, we do not enforce any privacy mechanism on the model.} (2) $L1$ (server-noise): in the case where the final model has to be sent back to the parties or even published, it should satisfy differential privacy guarantees to protect against potential inference attacks. (3) $L2$ (party-noise): in the case where the server is curious and all the models transferred from the parties to the server during training should satisfy differential privacy guarantees. \HBS{These three levels have their own application scenarios in practice. For example, in the model market, $L0$ is for a model market within an enterprise where different departments can share model but not the data due to data privacy regulations, and departments are trusted. $L1$ is similar to $L0$ but additionally the final model will be published in the market. $L2$ is for the public market where sellers need to protect their own data privacy.  }

\paragraph{The Overall Framework} The framework of FedKT is shown in Figure \ref{fig:framwork}. \HBS{It is designed to be a simple and practical one-shot algorithm for cross-silo settings.} Specifically, FedKT adopts a two-tier knowledge transfer structure. On the party side, each party uses knowledge transfer to learn student models and sends them to the server. On the server side, the server takes the received models as teachers to learn a final model using knowledge transfer again. The final model is sent back to the parties and used for predictions. Two techniques, multi-partitioning and consistent voting, are proposed to improve the performance.


\begin{algorithm}[t]
\caption{The FedKT algorithm}
\label{alg:fedkt}
\SetNoFillComment
\LinesNumbered
\KwIn{local datasets $\D^1, ..., \D^n$, number of partitions $s$ in each party, number of subsets $t$ in each partition, number of classes $u$, public dataset $\D_{aux}$, privacy parameter $\gamma$, privacy level $l$}
\KwOut{The final model $F$.}
\For {$i=1, ..., n$}{
\tcc{Conduct on party $P_i$}
    Create $s$ partitions (i.e., $D_1^i, ..., D_s^i$) on dataset $\D^i$ such that $D_j^i = \bigcup_{k\in[t]}D_{j,k}^i$ for all $j \in [s]$, where $D_{j,k}^i$ is a subset. \\
    \For {$j=1, ..., s$}{
        \For {$k=1, ..., t$}{
            Train a teacher model $T_{j,k}^i$ on subset $D_{j,k}^i$.
        }
        \For {all $\mb{x} \in \D_{aux}$}{
            \For {all $m \in [u]$}{
                $v_m(\mb{x}) \leftarrow |\{k: k\in [t], T_{j,k}^i(\mb{x}) = m\}|$\\
                \If {$l == L2$}{
                    $v_m(\mb{x}) \leftarrow v_m(\mb{x}) + Lap(1/\gamma)$
                }
            }
            $f(\mb{x}) = \argmax_{m} v_m(\mb{x})$
        }
        Train a student model $U_j^i$ on dataset $\{(\mb{x},f(\mb{x}))\}_{\mb{x}\in\D_{aux}}$.
    }
    Send the student models $\{U_j^i: j\in[s]\}$ to server.
}
\tcc{Conduct on the server}
\For {all $\mb{x} \in \D_{aux}$}{
    \For {all $i \in [n]$}{
        \For {all $m \in [u]$}{
            $v_m^i(\x) \leftarrow |\{k: k\in[s], U_k^i(\x)=m\}|$
        }
    }
    \For {all $m \in [u]$}{
        $v_m(\mb{x}) \leftarrow s\cdot|\{i: i\in[n], v_m^i(\x)=s\}|$\\
        \If {$l == L1$}{
            $v_m(\mb{x}) \leftarrow v_m(\mb{x}) + Lap(1/\gamma)$
        }
    }
    $f(\mb{x}) = \argmax_{m} v_m(\mb{x})$
}
Train the final model $F$ on dataset $\{(\mb{x},f(\mb{x}))\}_{\mb{x}\in\D_{aux}}$.
\end{algorithm}

\paragraph{Learning Student Models on the Parties (Multi-Partitioning)} Locally, each party has to create $s$ ($s \geq 1$) partitions and learn a student model on each partition. Each partition handles the entire local dataset. Since the operations in each partition are similar, here we describe the process in one partition for ease of presentation. Inside a partition, the local dataset is divided into $t$ disjoint subsets. We train a teacher model separately on each subset, denoted as $T_1, ..., T_t$. Then, the ensemble of teacher models is used to make predictions on the public dataset $\D_{aux}$. For an example $\x\in \D_{aux}$, the \emph{vote count} of class $m$ is the number of teachers that predicts $m$, i.e., $v_m(\x)=|\{i: i\in[t], T_i(\x)=m\}|$. The prediction result of the ensemble is the class that has the maximum vote counts, i.e., $f(\x)=\argmax_m v_m(\x)$. Then, we use the public dataset $\D_{aux}$ with the predicted labels to train a student model. For each partition, we get a student model with the above steps. After all the student models are trained, the parties send their student models to the server for further processing.

\paragraph{Learning the Final Model on the Server} Suppose the student models of party $i$ are denoted as $U_1^i, ..., U_s^i$ ($i \in [n]$). After receiving all the student models, like the steps on the party side, the server can use these student models as an ensemble to make predictions on the public dataset $\D_{aux}$. The public dataset with the predicted labels is used to train the final model. 

\paragraph{Consistent voting} Here we introduce a technique named \emph{consistent voting} for computing the vote counts of each class. If the student models of a party make the same prediction on an example, we take their predictions into account. Otherwise, the party is not confident at predicting this example and thus we ignore the predictions of its student models. Formally, given an example $\x\in\D_{aux}$, we first compute the vote count of class $m$ on the student models of party $i$ as $v_m^i(\x)=|\{k: k\in[s], U_k^i(\x)=m\}|$. Next, with consistent voting, the final vote count of class $m$ on all parties is computed as $v_m(\x) = s\cdot|\{i: i\in[n], v_m^i(\x)=s\}|$.

\paragraph{Differentially Private Versions of FedKT}
FedKT can easily satisfy differential privacy guarantees by providing differentially private prediction results to the query dataset. Given the privacy parameter $\gamma$, we can add noises to the vote count histogram such that $f(\x)=\argmax_m \{v_m(\x)+Lap(\frac{1}{\gamma})\}$, where $Lap(\frac{1}{\gamma})$ is the noises generated from Laplace distribution with location 0 and scale $\frac{1}{\gamma}$. Note that we do not need to add noises on both the parties and the server. For the $L1$ setting, we only need to add noises on the server side. The parties can train and send non-differentially private student models to the server. For the $L2$ setting, we only need to add noises on the party side so that the student models are differentially private. Then, the final model naturally satisfies differential privacy guarantees. More analysis on privacy loss will be presented in Section~\ref{sec:pri}.

Algorithm \ref{alg:fedkt} shows the whole training process of FedKT. In the algorithm, for each party (i.e., Line 1) and its each partition (i.e., Line 3), we train a student model using knowledge transfer (i.e., Lines 4-12). The student models are sent to the server. Then, the server trains the final model using knowledge transfer again (i.e., Lines 14-23). For different privacy level settings, we have the corresponding noises injection operations on the server side (i.e., Lines 20-21) or the party side (i.e., Lines 9-10).

\paragraph{Communication/Computation Overhead of FedKT} Suppose the size of each model is $M$. Then, the total communication size of FedKT is $nsM$ for sending the student models to the server. Suppose the number of communication rounds in FedAvg is $r$ and all the parties participate in the training in every iteration. Then the total communication size of FedAvg is $2nMr$ including the server sends the global model to the parties \footnote{It also happens in the first round to ensure all parties have the same initialized model \cite{mcmahan2016communication}.} and the parties send the local models to the server. Thus, when $r > \frac{s}{2}$, the communication cost of FedAvg is higher than FedKT. This value can be quite small, e.g., $r=2$ if we set $s=2$. Moreover, when $s=2$, FedAvg has the same communication cost with FedKT in the first round. The computation overhead of FedKT is usually larger than FedAvg in each round since FedKT needs to train multiple teacher and student models. However, the computation overhead is acceptable in the cross-silo setting, where the parties (e.g., companies, data centers) usually have a relatively large computation power.

\section{Data-Dependent Privacy Analysis of FedKT}
\label{sec:pri}
In this section, we use the moments accountant method \cite{abadi2016deep} to track the privacy loss in the training process. For $L1$ setting, we mainly consider the party-level differential privacy \cite{geyer2017differentially,brendan2018learning}, which is more attractive in the federated setting. Instead of aiming to protect a single record, party-level differential privacy ensures that the learned model does not reveal whether a party participated in federated learning or not. For $L2$ setting, we mainly consider the example-level differential privacy since it is not practical to require each local model to satisfy party-level differential privacy. For proofs of the theorems in this section, please refer to Section \ref{sec:pri} of Appendix.

\paragraph{FedKT-L1} Considering we change the whole dataset of a party, then at most $s$ student models will be influenced. Thus, on the server side, the sensitivity of the vote count histogram is $2s$ (i.e., the vote count of a class increases by $s$ and the vote count of another class decreases by $s$). According to the Laplace mechanism, we have the following theorem.  

\begin{theorem} Let $\M$ be the $f$ function executed on the server side. Given the number of partitions $s$ and the privacy parameter $\gamma$, $\M$ satisfies $(2s\gamma, 0)$ party-level differential privacy.
\label{thm:pldp}
\end{theorem}

Similar with \cite{papernot2016semi}, we conduct a data-dependent privacy analysis for FedKT with the moments accountant method.

\begin{theorem}
Let $\M$ be $(2s\gamma, 0)$ party-level differentially private. Let $q \geq \Pr[\M(d) \neq o^*]$ for some outcome $o^*$. Let $l,\gamma \geq 0$ and $q <\frac{e^{2s\gamma}-1}{e^{4s\gamma}-1}$. For any $\textsf{aux}$ and any two party-adjacent datasets $d$ and $d'$, $\M$ satisfies
	\begin{align*}
	\alpha_\M(l; \textsf{aux}, d, d') \leq 
	&\min(\log ((1-q)\Big(\frac{1-q}{1-e^{2s\gamma}q}\Big)^l + qe^{2s\gamma l}),\\
	&2 s^2 \gamma^2 l(l+1)).
	\end{align*}
\label{thm:mom}
\end{theorem}

Here $\Pr[\M(d) \neq o^*]$ can be bounded by Lemma 4 of \cite{papernot2016semi}.

With Theorem \ref{thm:mom}, we can track the privacy loss of each query \cite{abadi2016deep}.

\paragraph{FedKT-L2} For each partition on the party side, we add Laplace noises to the vote counts, which is same with the PATE approach. Thus, we have the following theorem.

\begin{theorem}
Let $\M$ be the $f$ function executed on each partition of a party. Let $q \geq \Pr[\M(d) \neq o^*]$ for some outcome $o^*$. Let $l,\gamma \geq 0$ and $q <\frac{e^{2\gamma}-1}{e^{4\gamma}-1}$. For any $\textsf{aux}$ and any two adjacent datasets $d$ and $d'$, $\M$ satisfies
	\begin{align*}
	\alpha_\M(l; \textsf{aux}, d, d') &\leq 
	\min(\log ((1-q)\Big(\frac{1-q}{1-e^{2\gamma}q}\Big)^l + qe^{2\gamma l}),
	\\&2 \gamma^2 l(l+1)).
	\end{align*}
\label{thm:mom2}
\end{theorem}

After computing the privacy loss of each party, we can use the parallel composition to compute the privacy loss of the final model.

\begin{theorem}
Suppose the student models of party $P_i$ satisfy $(\varepsilon_i, \delta)$-differential privacy. Then, the final model satisfies $(\max_i\varepsilon_i, \delta)$-differential privacy. 
\label{thm:para}
\end{theorem}

Note that the above privacy analysis is data-dependent. Thus, the final privacy budget is also data-dependent and may have potential privacy breaches if we publish the budget. Like previous studies \cite{papernot2016semi,jordon2019differentially}, we report the data-dependent privacy budgets in the experiments. As future work, we plan to use the smooth sensitivity algorithm \cite{nissim2007smooth} to add noises to the privacy losses.  Also, we may get a tighter bound of the privacy loss if adopting the Gaussian noises \cite{papernot2018scalable}. 
\section{Experiments}
To evaluate FedKT, we conduct experiments on four public datasets: (1) A random forest on \emph{Adult} dataset. The number of trees is set to 100 and the maximum tree depth is set to 6. (2) A gradient boosting decision tree (GBDT) model on \emph{cod-rna} dataset. The maximum tree depth is set to 6. (3) A multilayer perceptron (MLP) with two hidden layers on \emph{MNIST} dataset. Each hidden layer has 100 units using ReLu activations. (4) A CNN on extended \emph{SVHN} dataset. The CNN has two 5x5 convolution layers followed with 2x2 max pooling (the first with 6 channels and the second with 16 channels), two fully connected layers with ReLu activation (the first with 120 units and the second with 84 units), and a final softmax output layer. For the first two datasets, we split the original dataset at random into train/test/public sets with a 75\%/12.5\%/12.5\% proportion. For MNIST and SVHN, we use one half of the original test dataset as the public dataset and the remaining as the final test dataset. Like many existing studies \cite{pnfm,lin2020ensemble,li2021model}, we use the Dirichlet distribution to simulate the heterogeneous data partition among the parties. Suppose there are $n$ parties. We sample $p_k \sim Dir_n(\beta)$ and allocate a $p_{k,j}$ proportion of the instances of class $k$ to party $j$, where $Dir(\beta)$ is the Dirichlet distribution with a concentration parameter $\beta$ (0.5 by default). By default, we set the number of parties to 50 for Adult and cod-rna and to 10 for MNIST and SVHN. We set $s$ to 2 and $t$ to 5 by default for all datasets. For more details in the experimental settings and study of the hyper-parameters, please refer to Section \ref{sec:expdetail} and Section \ref{sec:sensi} of Appendix. The code is publicly available \footnote{\url{https://github.com/QinbinLi/FedKT}}.

We compare FedKT with eight baselines: (1) SOLO: each party trains its model locally and does not participate in federated learning. (2) FedAvg \cite{mcmahan2016communication}; (3) FedProx \cite{li2018federated}; (4) SCAFFOLD \cite{karimireddy2019scaffold}; (5) FedDF \cite{lin2020ensemble}; (6) PNFM \cite{pnfm};  (7) PATE \cite{papernot2016semi}: we use the PATE framework to train a final model on all data in a centralized setting (i.e., only a single party with the whole dataset) without adding noises. This method defines an upper bound of learning a final model using knowledge transfer with public unlabelled data. \del{We set the number of teacher models equals to the number of parties. }(8) XGBoost \cite{chen2016xgboost}: the XGBoost algorithm for the GBDT model on the whole dataset in a centralized setting. This method defines an upper bound of learning the GBDT model. Here approaches (2)-(5) are popular or state-of-the-art federated learning algorithms and approach (6) is an one-shot algorithm. Moreover, same as FedKT, approaches (5) and (7) also utilize the unlabeled public dataset.

\begin{table*}[t]
\centering
\caption{The test accuracy comparison between FedKT and the other baselines \HBS{in a single round}.}
\label{tbl:effectiveness}
\begin{tabular}{|c|c|c|c|c|c|c|c||c|c|}
\hline
Datasets & FedKT & SOLO & FedAvg & FedProx & SCAFFOLD & FedDF & PNFM & PATE & XGBOOST \\ \hline
Adult & \tb{82.2\%} $\pm$ 0.6\%& 68.6\% & \multicolumn{5}{c||}{\multirow{2}{*}{\backslashbox{}{}}} & 83.5\% & \backslashbox{}{} \\ \cline{1-3} \cline{9-10} 
cod-rna & \tb{88.3\%} $\pm$ 0.6\% & 65.0\% & \multicolumn{5}{c||}{} & 91.1\% & 91.2\% \\ \hline
MNIST & \tb{90.5\%} $\pm$ 0.3\%& 69.0\% & 62.8\% & 44.3\% & 51.7\% & 83.8\% & 65.9\% & 92.7\% &\backslashbox{}{}  \\ \hline
SVHN & \tb{83.2\%} $\pm$ 0.4\%& 62.8\% & 26.8\% & 20.1\% & 16.2\% & 77.2\% & \backslashbox{}{} & 86.6\% & \backslashbox{}{} \\ \hline
\end{tabular}
\end{table*}

\begin{table*}[t]
\centering
\caption{The privacy loss and test accuracy of FedKT-L1 and FedKT-L2 given different $\gamma$ and number of queries. L0 acc is the test accuracy of FedKT-L0. The failure probability $\delta$ is set to $10^{-5}$.}
\label{tbl:dp}
\resizebox{0.98\textwidth}{!}{%
\begin{tabular}{|c|c|c|c|c|c||c|c|c|c|c|}
\hline
\multirow{2}{*}{datasets} & \multicolumn{5}{c||}{FedKT-L1} & \multicolumn{5}{c|}{FedKT-L2}\\ \cline{2-11}
 & $\gamma$ & \#queries & $\varepsilon$ & acc & non-private acc   & $\gamma$ & \#queries &  $\varepsilon$ & acc & non-private acc   \\ \hline
\multirow{2}{*}{Adult} & 0.04 & 0.5\% & 2.56 & 76.8\% & \multirow{2}{*}{82.2\%} & 0.04 & 0.5\% & 2.59 & 77.6\% & \multirow{2}{*}{82.4\%} \\ \cline{2-5}\cline{7-10}
 & 0.04 & 1.0\% & 4.73 & 80.2\% &    & 0.04 & 1.0\% & 3.72 & 78.6\% &    \\ \cline{1-11}
\multirow{2}{*}{cod-rna} & 0.06 & 0.50\% & 5.48 & 82.6\% & \multirow{2}{*}{88.0\%}   & 0.05 & 1.0\% & 4.51 & 82.7\% & \multirow{2}{*}{89.7\%}   \\ \cline{2-5}\cline{7-10}
 & 0.1 & 0.5\% & 6.89 & 84.7\% &    & 0.05 & 2.0\% & 9.78 & 84.7\% &    \\ \hline
\end{tabular}
}
\end{table*}

\subsection{Effectiveness}

Table \ref{tbl:effectiveness} shows the accuracy of FedKT\footnote{For simplicity, we use FedKT to denote FedKT-L0, unless specified otherwise.} compared with the other baselines. For fair comparison \HBS{and practical usage in model markets}, we run all approaches for a single communication round. For SOLO, we report the average accuracy of the parties. From this table, we have the following observations. First, except for FedKT, the other federated learning algorithms can only learn specific models. FedKT is able to learn all the studied models including trees and neural networks. Second, FedKT can achieve much better performance than the other federated learning algorithms running with a single round. FedKT can achieve about 6.5\% higher accuracy than FedDF, which also utilizes the public dataset. Third, although PNFM is an one-shot algorithm specially designed for MLPs, FedKT outperforms PNFM about 25\% accuracy on MNIST. Last, the accuracy of FedKT is close to PATE and XGBoost, which means our design has little accuracy loss compared with the centralized setting. The gap between FedKT and the upper bound is very small.

\subsection{Privacy}
We run FedKT with different $\gamma$ and number of queries. When running FedKT-L1, we tune the percentage of number of queries on the server side. When running FedKT-L2, on the contrary, we tune the percentage of number of queries on the party side. The selected results on Adult and cod-rna are reported in Table \ref{tbl:dp}. While differentially private FedKT does not need any knowledge on the model architecture, the accuracy is still comparable to the non-private version given a privacy budget less than 10. For more results, please refer to Section \ref{sec:dpexp} of Appendix.

\subsection{\rev{Size of the Public Dataset}}
\label{sec:pub_size}
\rev{We show the performance of FedKT with different size of public dataset. The results are shown in Table \ref{tbl:public_size}. We can observe that FedKT is stable even though reducing the size of the public dataset. The accuracy decreases no more than 1\% and 2\% using only 20\% of the public dataset on cod-rna and MNIST (i.e., 1807 examples on cod-rna and 1000 examples on MNIST), respectively. Moreover, the accuracy is almost unchanged on Adult.}

\begin{table}[]
\centering
\caption{\rev{The test accuracy of FedKT with different size of the public dataset. We vary the portion of the public dataset used in the training from 20\% to 100\%.}}
\label{tbl:public_size}
\begin{tabular}{|c|c|c|c|c|c|}
\hline
\multirow{2}{*}{Datasets} & \multicolumn{5}{c|}{Portion of the public dataset used in training} \\ \cline{2-6} 
 & 20\% & 40\% & 60\% & 80\% & 100\% \\ \hline\hline
Adult & 82.1\% & 82.1\% & 82.1\% & 82.3\% & 82.2\% \\ \hline
cod-rna & 87.3\% & 87.6\% & 87.8\% & 87.8\% & 88.3\% \\ \hline
MNIST &89.3\% &89.7\% &90.2\% &90.3\% & 90.5\% \\ \hline
SVHN & 80.1\% & 81.3\% & 82.1\% &82.9\% & 83.2\% \\ \hline
\end{tabular}
\end{table}

\subsection{Extend to Multiple Rounds}

While FedKT is an one-shot algorithm, it is still applicable in the scenarios where multiple rounds are allowed. FedKT can be used as an initialization step to learn a global model in the first round. Then the parties can use the global model to conduct iterative federated learning algorithms. By combining FedKT with FedProx (denoted as FedKT-Prox), FedKT-Prox is much more communication-efficient than the other algorithms. FedKT-Prox needs about 11 rounds to achieve 87\% accuracy, while the other approaches need at least 32 rounds. For more details, please refer to Section \ref{sec:multi_round} of Appendix.

\section{Conclusions}
\HBS{Motivated by the rigid multi-round training of current federated learning algorithms and emerging applications like model markets, we propose FedKT, a one-shot federated learning algorithm for the cross-silo setting.} Our experiments show that FedKT can learn different models with a much better accuracy then the other state-of-the-art algorithms with a single communication round. Moreover, the accuracy of differentially private FedKT is comparable to the non-differentially private version with a modest privacy budget. Overall, \HBS{FedKT is a practical one-shot solution for model-based sharing in cross-silo federated learning.}

\section*{Acknowledgements}
This research is supported by the National Research Foundation, Singapore under its AI Singapore Programme (AISG Award No: AISG2-RP-2020-018). Any opinions, findings and conclusions or recommendations expressed in this material are those of the authors and do not reflect the views of National Research Foundation, Singapore.
\bibliographystyle{named}
\bibliography{ijcai21}

\newpage
\appendix
\section*{Appendix}

In this appendix, we first present the data-dependent privacy analysis of FedKT-L1 and FedKT-L2 in Section \ref{sec:priv}. Next, in Section \ref{sec:expdetail}, we show the details of our experimental settings. Then, we show additional experimental results in Section \ref{sec:sensi} to Section \ref{sec:dpexp}. Specifically, in Section \ref{sec:sensi}, we study the performance of FedKT with different hyper-parameters. In Section \ref{sec:pub_size}, we study the effect of the size of the public dataset. In Section \ref{sec:multi_round}, we extend FedKT to multiple rounds. Last, in section \ref{sec:dpexp}, we show the performance of FedKT-L1 and FedKT-L2.

\section{Privacy Analysis of FedKT}
\label{sec:priv}
In this section, we analyze the privacy loss of FedKT-L1 and FedKT-L2 using the moments accountant method \cite{abadi2016deep}.

\subsection{Proof of Theorem 2}
\label{sec:proof2}

\begin{proof}
We first introduce two theorems from the previous studies, which will be used in our analysis. The first theorem is from \cite{bun2016concentrated} and the second theorem is from \cite{papernot2016semi}.

\begin{lemma}
Let $\M$ be $(2\gamma, 0)$-differentially private. For any $l$, $\aux$, neighboring inputs $d$ and $d'$, we have
	\begin{align*}
	\alpha_\M(l; \textsf{aux}, d, d') \leq 2 \gamma^2 l(l+1)
	\end{align*}
\label{thm:pre1}
\end{lemma}

\begin{lemma}
Let $\M$ be $(2\gamma, 0)$-differentially private and $q \geq \Pr[\M(d) \neq o^*]$ for some outcome $o^*$. Let $l,\gamma \geq 0$ and $q <\frac{e^{2\gamma}-1}{e^{4\gamma}-1}$. Then for any $\textsf{aux}$ and any neighbor $d'$ of $d$, $\M$ satisfies
	\begin{align*}
	\alpha_\M(l; \textsf{aux}, d, d') \leq 
	\log ((1-q)\Big(\frac{1-q}{1-e^{2\gamma}q}\Big)^l + qe^{2\gamma l})
	\end{align*}
\label{thm:pre2}
\end{lemma}

$\Pr[\M(d) \neq o^*]$ can be bounded by the following lemma.

\begin{lemma}
  Let $\mathbf{v}$ be the label score vector for an instance $d$ with $v_{o^*} \geq v_o$ for all $o$. Then
  \begin{align*}
    \Pr[\M(d) \neq o^*] \leq \sum_{o \neq o^*} \frac{2 + \gamma(v_{o^*} - v_o)}{4 \exp(\gamma(v_{o^*} - v_o))}
  \end{align*}
  \label{lem:1}
\end{lemma}

According to Theorem 1 of the paper, $\M$ satisfies (2$s\gamma$, 0) party-level differential privacy. Thus, substituting $s\gamma$ into Lemma \ref{thm:pre1} and \ref{thm:pre2}, we can get 

	\begin{align*}
	\alpha_\M(l; \textsf{aux}, d, d') \leq 
	&\min(\log ((1-q)\Big(\frac{1-q}{1-e^{2s\gamma}q}\Big)^l + qe^{2s\gamma l}),\\
	&2 s^2 \gamma^2 l(l+1)).
	\end{align*}

\end{proof}

\subsection{Proof of Theorem 3}
\begin{proof}
The noises injection on the party side is similar to the PATE approach. The sensitivity of the $f$ function executed on the party side is 2. Thus, we can get the theorem by combining Lemma \ref{thm:pre1} and Lemma \ref{thm:pre2}. We have 
	\begin{align*}
	\alpha_\M(l; \textsf{aux}, d, d') &\leq 
	\min(\log ((1-q)\Big(\frac{1-q}{1-e^{2\gamma}q}\Big)^l + qe^{2\gamma l}),
	\\&2 \gamma^2 l(l+1)).
	\end{align*}
\end{proof}

\subsection{Example-Level Differential Privacy Analysis of FedKT-L1}

Here we analyze the example-level differential privacy of FedKT-L1. If we change a single example of the original dataset (i.e., the union of all the local datasets), only a single party will be influenced. More precisely, for each partition of the party, only a single teacher model will be influenced. Then, even though changing a single record, the student model is still unchanged if the top-2 vote counts of the teachers differ at least 2. Thus, if not applying consistent voting, we have the following theorem.

\begin{theorem}
\label{thm:l1record}
Let $\M$ be the $f$ function executed in the server. Let $q\geq \Pr[\M(d)\neq o^*]$ for some outcome $o^*$. Let $\D_{aux}$ denotes the query dataset.
Given a query $i$, suppose the top-2 vote counts are $v_1^i$ and $v_2^i$. In party $P_i$, let $z_i$ denotes the number of partitions that there $\exists q \in \D_{aux}$ such that $v_1^q-v_2^q \leq 1$ when training the student model. Let $z=max_i z_i$. Let $l,\gamma \geq 0$ and $q <\frac{e^{2z\gamma}-1}{e^{4z\gamma}-1}$. We have

\begin{align*}
  \alpha_\M(l; \textsf{aux},d,d') \leq &\min(\log ((1-q)\Big(\frac{1-q}{1-e^{2z\gamma}q}\Big)^l + qe^{2z\gamma l}),\\ &2z^2\gamma^2l(l+1))
\label{eq:mom2}
\end{align*}

\end{theorem}

\begin{proof}
Given a query dataset $\D_{aux}$, $z$ is the number of partitions such that there exists a query that the top-2 vote counts differ at most 1. In other words, there are at most $z$ student models will be changed if we change a single record of the original dataset. Thus, the vote counts change by at most $2z$ on the server side and $\M$ is $(2z\gamma, 0)$-differentially private with respect to $d$ and $\D_{aux}$. Then, we can get this theorem by substituting $\gamma$ of Theorem \ref{thm:pre1} and Theorem \ref{thm:pre2} to $z\gamma$.
\end{proof}

Note that the example-level differential privacy is same as party-level differential privacy when $z=s$. Also, if we applied consistent voting in FedKT-L1, the vote counts may change by $s$ even if only a single student model is affected. Thus, the example-level differential privacy of FedKT-L1 is usually same as party-level differential privacy. 

\subsection{Party-Level Differential Privacy Analysis of FedKT-L2}
\label{sec:fedktl2}

Here we study party-level differential privacy of FedKT-L2.

\begin{theorem} Let $\M$ be the $f$ function executed on each partition of a party. Given the number of subsets in each partition $t$ and the privacy parameter $\gamma$, $\M$ satisfies $(2t\gamma, 0)$ party-level differential privacy.
\label{thm:pldpl2}
\end{theorem}

\begin{proof}
Considering changing the whole local dataset, then $t$ teachers will be influenced and the vote counts change by at most $2t$. Thus, from a party-level perspective, the sensitivity of the vote counts is $2t$ and $\M$ satisfies $(2t\gamma, 0)$-differential privacy. 
\end{proof}

Like Theorem \ref{thm:l1record}, we have the following theorem to track the moments.

\begin{theorem}
Let $\M$ be the $f$ function executed on the party side and $q \geq \Pr[\M(d) \neq o^*]$ for some outcome $o^*$. Suppose the number of subsets in a partition is $t$. Let $l,\gamma \geq 0$ and $q <\frac{e^{2t\gamma}-1}{e^{4t\gamma}-1}$. Then for any $\textsf{aux}$ and any two party-adjacent datasets $d$ and $d'$, $\M$ satisfies
	\begin{align*}
	\alpha_\M(l; \textsf{aux}, d, d') \leq 
	&\min(\log ((1-q)\Big(\frac{1-q}{1-e^{2t\gamma}q}\Big)^l + \\&qe^{2t\gamma l}), 2 t^2\gamma^2 l(l+1)).
	\end{align*}
\label{thm:l2party}
\end{theorem}

Note that the party-level privacy loss of FedKT-L2 can be quite large. To mitigate the impact of the noises, we usually expect $t$ to be large to have a tighter bound of $\Pr[\M(d) \neq o^*]$. However, when $t$ is large, the bound in Theorem \ref{thm:l2party} is also large. In fact, since every student model satisfies differential privacy, it is not necessary to apply party-level differential privacy in FedKT-L2.

\section{Experiments}
\label{sec:exp_appendix}

\subsection{Additional Details of Experimental Settings}
\label{sec:expdetail}

We use \emph{Adult}, \emph{cod-rna}, \emph{MNIST}, and \emph{SVHN} for our experiments, where \emph{Adult} and \emph{cod-rna} are downloaded from this link\footnote{\url{https://www.csie.ntu.edu.tw/~cjlin/libsvmtools/datasets/}}. The details of the datasets are shown in Table \ref{tbl:data}. We run experiments on a Linux cluster with 8 RTX 2080 Ti GPUs and 12 Intel Xeon W-2133 CPUs.

\begin{table*}[t]
\centering
\caption{The datasets used and their learning models. The detailed model structures are shown in the paper.}
\label{tbl:data}
\begin{tabular}{|c|c|c|c|c|}
\hline
Datasets & Adult & cod-rna & MNIST &SVHN \\ \hline \hline
\#Training examples & 24421 & 54231 & 50000 & 604288\\ \hline
\#Public examples & 4070 & 9039 & 5000 &13016 \\ \hline
\#Test examples & 4070 & 9039 & 5000 &13016 \\ \hline
\#Classes & 2 & 2 & 10 &10\\ \hline
\#Parties (by default) & 50 & 50 & 10 &10 \\ \hline
Model & Random Forest & GBDT & MLP & CNN\\ \hline
Implemented library & scikit-learn 0.22.1 & XGBoost 1.0.2 & PyTorch 1.6.0 &PyTorch 1.6.0\\ \hline
\end{tabular}
\end{table*}

We compare FedKT with the other eight baselines. For the federated learning algorithms, the learning rate is tuned from $\{0.001, 0.01\}$ and the number of local epochs is tuned from $\{10, 20, 40\}$. We found that the baselines can get best performance when the learning rate is set to 0.001 and the number of local epochs is set to 10. For FedProx, the regularization term $\mu$ is tuned from $\{0.1, 1\}$. For SCAFFOLD, same as the experiments of the paper \cite{karimireddy2019scaffold}, we use option \rom{2} to calculate the control variates. For FedKT, SOLO, and PATE, the number of local epochs is simply set to 100. For PATE, the number of teacher models is set to be same as the number of parties of FedKT. We use the Adam optimizer and the $L_2$ regularization is set to $10^{-6}$. The final parameters are summarized in Table \ref{tbl:para}.

\begin{table*}[t]
\centering
\caption{The default parameters used in our experiments.}
\label{tbl:para}
\begin{tabular}{|c|c|c|c|c|c|}
\hline
\multicolumn{2}{|c|}{Parameters} & Adult & cod-rna & MNIST & SVHN \\ \hline\hline
\multirow{5}{*}{common} & \#parties & 50 & 50 & 10 & 10 \\ \cline{2-6} 
 & tree depth & 6 & 6 & \backslashbox{}{} & \backslashbox{}{} \\ \cline{2-6} 
 & learning rate &\backslashbox{}{}  & 0.05 & 0.001 & 0.001 \\ \cline{2-6} 
 & batch size & \backslashbox{}{} &\backslashbox{}{}  & 32 & 64 \\ \cline{2-6} 
 & \#epochs & \backslashbox{}{} &\backslashbox{}{}  & 10 & 10 \\ \hline
\multirow{2}{*}{FedKT} & number of partitions in a party & 2 & 2 & 2 & 2 \\ \cline{2-6} 
 & number of subsets in a partition & 5 & 5 & 5 & 5 \\ \hline
FedProx & regularization term $\mu$ &\backslashbox{}{}  &\backslashbox{}{} & 0.1 & 0.1 \\ \hline
\end{tabular}
\end{table*}

\subsection{Hyper-parameters Study}
\label{sec:sensi}

\subsubsection{Number of partitions in each party}
Here we study the impact of the number of partitions (i.e., the parameter $s$) on FedKT. Table \ref{tbl:tunepartition} shows the test accuracy of FedKT with different $s$. From Table \ref{tbl:tunepartition}, we can see that the accuracy can be improved if we increase $s$ from 1 to 2. However, if we further increase $s$, there is little or no improvement on the accuracy while the communication and computation overhead is larger. Thus, from our empirical study, we suggest users to simply set $s$ to 2 for FedKT-L0 if they do not want to tune the parameters. For FedKT-L1 and FedKT-L2, since the privacy loss increases as $s$ increases, we suggest to set $s$ to small values. Users can simply set $s$ to 1 or tune $s$ from small values (i.e., 1 or 2) to find the best accuracy-privacy trade-off.

\begin{table*}[!]
\centering
\caption{The test accuracy of FedKT with number of partitions ranging between 1 and 5. We run 5 trials and report the mean and standard deviation. The number of subsets in each partition is set to 5 by default.}
\label{tbl:tunepartition}
\begin{tabular}{|c|c|c|c|c|c|}
\hline
\#partitions & 1 & 2 & 3 & 4 &5 \\ \hline\hline
Adult &$80.8\%\pm1.4\%$ &$\tb{82.2\%}\pm0.6\%$ &$81.5\%\pm0.6\%$  &$81.2\%\pm0.5\%$ &$81.1\%\pm0.1\%$ \\ \hline
cod-rna & $87.7\%\pm0.6\%$ & $\tb{88.3\%}\pm0.6\%$& $\tb{88.3\%}\pm0.5\%$ &$88.1\%\pm0.5\%$ &$88.2\%\pm0.5\%$ \\ \hline
MNIST &$89.2\% \pm 0.4\% $&$\tb{90.5\%} \pm 0.3\%$ &$89.9\% \pm 0.2\%$ &$90.1\% \pm 0.2\%$ &$90.2\% \pm 0.2\%$  \\ \hline
SVHN &$81.5\% \pm 0.6\%$ & $83.2\%\pm0.4\%$ &$\tb{83.5\%}\pm0.4\%$ &$83.5\%\pm0.4\%$ &$83.4\%\pm0.3\%$\\ \hline
\end{tabular}
\end{table*}

\subsubsection{Number of teachers in each partition}

Here we study the impact of number of teachers in each partition (i.e., the parameter $t$). Table \ref{tbl:tunesubsets} shows the test accuracy of FedKT with different $t$. As we can see, FedKT can always get the best performance if setting $t$ to 5. If $t$ is large, the size of each data subset is small and the teacher models may not be good at predicting the public dataset. From our empirical study, users can simply set $t$ to 5 if they do not want to tune the parameter. However, if the student models need to satisfy differential privacy (i.e., in FedKT-L2), the privacy loss may potentially be smaller if we increase $t$ according to Lemma \ref{lem:1}. Users need to tune $t$ to find the best trade-off between the performance and the privacy loss. 

\begin{table}[]
\centering
\caption{The test accuracy of FedKT with number of subsets in each partition ranging between 5 and 20. We run 5 trials and report the mean and standard deviation. For Adult, since there is a party with less than 15 examples, the experiment cannot successfully run when the number of subsets is not smaller than than 15. For cod-rna, since there is a party with less than 20 examples, the experiment cannot successfully run when setting the number of subsets to 20. The number of partitions in each party is set to 2 by default.}
\label{tbl:tunesubsets}
\resizebox{\columnwidth}{!}{%
\begin{tabular}{|c|c|c|c|c|}
\hline
\#subsets & 5 & 10 & 15 & 20 \\ \hline\hline
Adult & $\tb{82.0\%}\pm0.6\%$  &$81.1\%\pm0.7\%$  & \backslashbox{}{} & \backslashbox{}{} \\ \hline
cod-rna & $\tb{88.3\%}\pm0.6\%$ &$87.4\%\pm0.6\%$  &$83.1\%\pm0.6\%$ & \backslashbox{}{} \\ \hline
MNIST &$\tb{90.5\%}\pm0.3\%$ &$89.4\% \pm 0.2\%$  &$89\% \pm 0.4\%$  & $88.4\% \pm 0.3\%$\\ \hline
SVHN &$\tb{83.2\%}\pm0.4\%$  &$81.3\%\pm0.5\%$  &$80.0\%\pm0.5\%$ &$79.1\%\pm0.6\%$  \\ \hline
\end{tabular}
}
\end{table}

\subsection{Extend to Multiple Rounds}
\label{sec:multi_round}
\begin{figure}[t]
\centering
\includegraphics[width=0.45\textwidth]{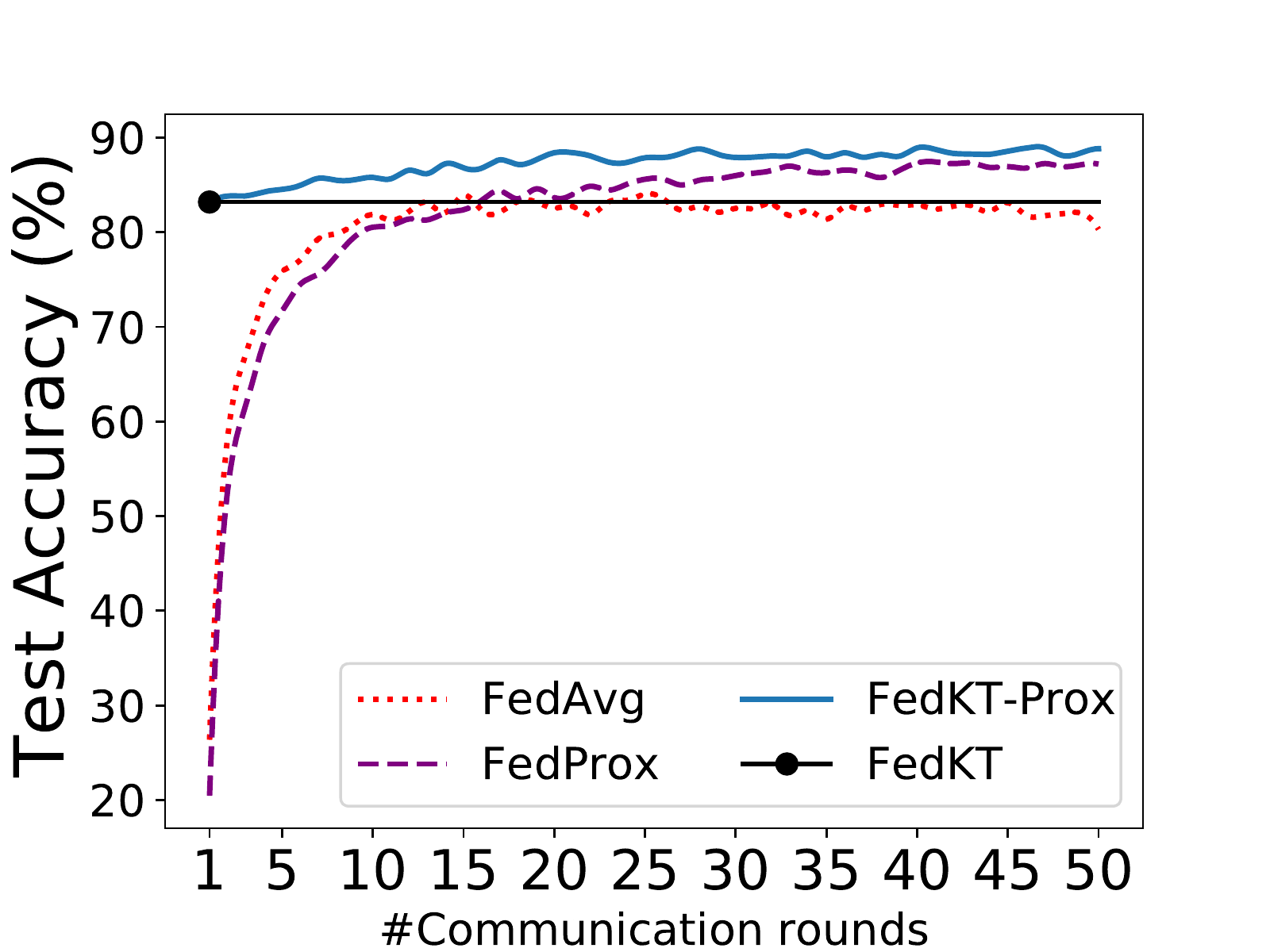}%
\caption{The test accuracy with increasing communication rounds/communication size on SVHN.}
\label{fig:comm_eff}
\end{figure}

FedKT can be used as an initialization step if applied to iterative training process. Here we combine FedKT and FedProx (denoted as FedKT-Prox) and compare it with FedAvg and FedProx. The results are shown in Figure \ref{fig:comm_eff}. We can observe that FedKT-Prox can always achieve much higher accuracy than FedAvg and FedProx. Overall, FedKT-Prox is much more communication-efficient than FedAvg and FedProx.

\subsection{Differential Privacy}
\label{sec:dpexp}
Table \ref{tbl:ktl1} and Table \ref{tbl:fedktl2} present the results of FedKT-L1 and FedKT-L2. Besides the heterogeneous partition ($\beta=0.5$), we also try homogeneous partition (i.e., the dataset is randomly and equally partitioned into the parties). From the tables, we can see that the accuracy of FedKT-L1 and FedKT-L2 are comparable to the non-private version with a modest privacy budget. Moreover, the moments accountant method usually can achieve a tighter privacy loss than the advanced composition \cite{dwork2014algorithmic}. For example, if we run cod-rna under homogeneous data partition setting $\gamma=0.1$ and the fraction of queries to 1\%, the advanced composition gives us $\varepsilon \approx 20.2$ and our analysis gives $\varepsilon \approx 11.2$.
Note that the techniques in \cite{papernot2018scalable} can also be applied to FedKT. For example, we may get a smaller privacy loss if adopting Gaussian noises instead of Laplace noises. Generally, our framework can also benefit from the state-of-the-art approaches on the privacy analysis of PATE, which we may investigate in the future.

\begin{table*}[]
\centering
\caption{The accuracy and party-level $\varepsilon$ of FedKT-L1 on Adult and cod-rna given different $\gamma$ values and number of queries. For each setting, we run 3 trials and report the median accuracy and the corresponding $\varepsilon$. The number of partitions is set to 1 and the number of subsets in each partition is set to 5. The failure probability $\delta$ is set to $10^{-5}$.}
\label{tbl:ktl1}
\begin{tabular}{|c|c|c|c|c|c|c|}
\hline
 & data partitioning & $\gamma$ & \#queries & $\varepsilon$ & acc & FedKT-L0 acc \\ \hline \hline
\multirow{18}{*}{Adult} & \multirow{9}{*}{Heterogeneous} & \multirow{3}{*}{0.04} & 0.1\% & 0.64 & 71.3\% & \multirow{9}{*}{82.2\%} \\ \cline{4-6}
 &  &  & 0.5\% & 2.56 & 76.8\% &  \\ \cline{4-6}
 &  &  & 1\% & \tb{4.73} & \tb{80.2\%} &  \\ \cline{3-6}
 &  & \multirow{3}{*}{0.06} & 0.1\% & 0.96 & 75.7\% &  \\ \cline{4-6}
 &  &  & 0.5\% & 3.64 & 77.6\% &  \\ \cline{4-6}
 &  &  & 1\% & 5.78 & 80.2\% &  \\ \cline{3-6}
 &  & \multirow{3}{*}{0.08} & 0.1\% & 1.23 & 76.0\% &  \\ \cline{4-6}
 &  &  & 0.5\% & 4.25 & 76.3\% &  \\ \cline{4-6}
 &  &  & 1\% & 7 & 80.3\% &  \\ \cline{2-7} 
 & \multirow{9}{*}{Homogeneous} & \multirow{3}{*}{0.02} & 0.1\% & 0.32 & 72.2\% & \multirow{9}{*}{82.4\%} \\ \cline{4-6}
 &  &  & 0.5\% & 1.25 & 76.1\% &  \\ \cline{4-6}
 &  &  & 1\% & 1.79 & 80.1\% &  \\ \cline{3-6}
 &  & \multirow{3}{*}{0.04} & 0.1\% & 0.6 & 76.0\% &  \\ \cline{4-6}
 &  &  & 0.5\% & 1.9 & 80.4\% &  \\ \cline{4-6}
 &  &  & 1\% & 3.32 & 81.5\% &  \\ \cline{3-6}
 &  & \multirow{3}{*}{0.06} & 0.1\% & 0.75 & 76.1\% &  \\ \cline{4-6}
 &  &  & 0.5\% & 2.03 & 81.7\% &  \\ \cline{4-6}
 &  &  & 1\% & \tb{3.36} & \tb{82.1\%} &  \\ \hline \hline
\multirow{18}{*}{cod-rna} & \multirow{9}{*}{Heterogeneous} & \multirow{3}{*}{0.04} & 0.1\% & 1.09 & 66.8\% & \multirow{9}{*}{88.3\%} \\ \cline{4-6}
 &  &  & 0.5\% & 3.54 & 72.5\% &  \\ \cline{4-6}
 &  &  & 1\% & 5.14 & 75.2\% &  \\ \cline{3-6}
 &  & \multirow{3}{*}{0.06} & 0.1\% & 1.52 & 69\% &  \\ \cline{4-6}
 &  &  & 0.5\% & 5.48 & 82.6\% &  \\ \cline{4-6}
 &  &  & 1\% & 8.1 & 79.5\% &  \\ \cline{3-6}
 &  & \multirow{3}{*}{0.1} & 0.1\% & 2.12 & 69\% &  \\ \cline{4-6}
 &  &  & 0.5\% & \tb{6.89} & \tb{84.7\%} &  \\ \cline{4-6}
 &  &  & 1\% & 11.2 & 85.3\% &  \\ \cline{2-7} 
 & \multirow{9}{*}{Homogeneous}  & \multirow{3}{*}{0.02} & 0.2\% & 0.53 & 67.0\% & \multirow{9}{*}{88.6\%} \\ \cline{4-6}
 &  &  & 0.5\% & 1.71 & 73.2\% &  \\ \cline{4-6}
 &  &  & 1\% & 2.45 & 75.3\% &  \\ \cline{3-6}
 &  & \multirow{3}{*}{0.04} & 0.2\% & 1.5 & 73\% & \\ \cline{4-6}
 &  &  & 0.5\% & 3.06 & 84.1\% &  \\ \cline{4-6}
 &  &  & 1\% & 5.10 & 85\% &  \\ \cline{3-6}
 &  & \multirow{3}{*}{0.06} & 0.2\% & 1.63 & 80.2\% &  \\ \cline{4-6}
 &  &  & 0.5\% & 3.10 & 84.1\% &  \\ \cline{4-6}
 &  &  & 1\% & \tb{5.14} & \tb{86.1\%} &  \\ \hline

\end{tabular}
\end{table*}

\begin{table*}[]
\centering\caption{The accuracy and example-level $\varepsilon$ of FedKT-L2 on Adult and cod-rna given different $\gamma$ values and number of queries. For each setting, we run 3 trials and report the median accuracy and the corresponding $\varepsilon$. The number of parties is set to 20 to ensure that FedKT has enough data to train each teacher model. The number of partitions is set to 1 and the number of subsets in each partition is set to 25. The failure probability $\delta$ is set to $10^{-5}$.}
\label{tbl:fedktl2}
\begin{tabular}{|c|c|c|c|c|c|c|}
\hline
 & Data Partition & $\gamma$ & \#queries & $\varepsilon$ & acc & FedKT-L0 acc \\ \hline \hline
\multirow{18}{*}{Adult} & \multirow{9}{*}{Heterogeneous} & \multirow{3}{*}{0.04} & 0.1\% & 1.13 & 76.1\% & \multirow{9}{*}{82.4\%} \\ \cline{4-6}
 &  &  & 0.5\% & 2.56 & 76.5\% &  \\ \cline{4-6}
 &  &  & 1\% & 3.72 & 78.5\% &  \\ \cline{3-6}
 &  & \multirow{3}{*}{0.05} & 0.1\% & 1.32 & 76.1\% &  \\ \cline{4-6}
 &  &  & 0.5\% & 3.24 & 79.0\% &  \\ \cline{4-6}
 &  &  & 1\% & 4.76 & 79.2\% &  \\  \cline{3-6}
  &  & \multirow{3}{*}{0.06} & 0.1\% &1.96 &76.2\%  &  \\ \cline{4-6}
 &  &  & 0.5\% &3.93 &78.5\% &  \\ \cline{4-6}
 &  &  & 1\% & \tb{5.79} & \tb{79.4\%} &  \\ \cline{2-7} 
 & \multirow{9}{*}{Homogeneous} & \multirow{3}{*}{0.04} & 0.3\% & 2.13 & 76.1\% & \multirow{9}{*}{82.6\%} \\ \cline{4-6}
 &  &  & 0.5\% & 2.59 & 78.7\% &  \\ \cline{4-6}
 &  &  & 1\% & \tb{3.72} & \tb{81.7\%} &  \\ \cline{3-6}
 &  & \multirow{3}{*}{0.06} & 0.3\% & 2.97 & 76.3\% &  \\ \cline{4-6}
 &  &  & 0.5\% & 3.93 & 79.9\% &  \\ \cline{4-6}
 &  &  & 1\% & 5.79 & 81.8\% &  \\ \cline{3-6}
  &  & \multirow{3}{*}{0.08} & 0.3\% & 3.77 & 76.3\% &  \\ \cline{4-6}
 &  &  & 0.5\% & 5.04 & 80.4\% &  \\ \cline{4-6}
 &  &  & 1\% & 7.89 & 82.0\% &  \\ \hline \hline
\multirow{18}{*}{cod-rna} & \multirow{9}{*}{Heterogeneous} & \multirow{3}{*}{0.04} & 0.5\% &3.54 &77.7\% & \multirow{9}{*}{89.7\%} \\ \cline{4-6}
 &  &  & 1\% &5.14  &79.8\%  &  \\ \cline{4-6}
 &  &  & 2\% &7.63  &82.0\%  &  \\ \cline{3-6}
 &  & \multirow{3}{*}{0.05} & 0.5\% &4.51  &81.4\%  &  \\ \cline{4-6}
 &  &  & 1\% &6.58 &82.0\% &  \\ \cline{4-6}
 &  &  & 2\% &\tb{9.78}  &\tb{84.7\%} &  \\ \cline{3-6}
  &  & \multirow{3}{*}{0.06} & 0.5\% &5.50  &81.2\%  &  \\ \cline{4-6}
 &  &  & 1\% &8.10 &83.2\% &  \\ \cline{4-6}
 &  &  & 2\% &12.2  &85.9\% &  \\ \cline{2-7}
 & \multirow{9}{*}{Homogeneous} & \multirow{3}{*}{0.03} & 0.5\% &2.64 &79.2\%  & \multirow{9}{*}{90.6\%} \\ \cline{4-6}
 &  &  & 1\% &3.78  &80.5\%  &  \\ \cline{4-6}
 &  &  & 2\% &5.51  &83.1\%  &  \\ \cline{3-6}
 &  & \multirow{3}{*}{0.04} & 0.5\% &3.54 &80.1\%  &  \\ \cline{4-6}
 &  &  & 1\% &5.14  &83.7\%  &  \\ \cline{4-6}
 &  &  & 1.5\% &6.45 &84.0\%  &  \\ \cline{3-6}
  &  & \multirow{3}{*}{0.05} & 0.5\% &4.51 &80.8\%  &  \\ \cline{4-6}
 &  &  & 1\% &6.58  &84.2\%  &  \\ \cline{4-6}
 &  &  & 1.5\% &\tb{8.3} &\tb{84.7\%}  &  \\ \hline
\end{tabular}
\end{table*}

\end{document}